\documentclass[conference]{IEEEtran}
\IEEEoverridecommandlockouts
\usepackage{cite}
\usepackage{amsmath,amssymb,amsfonts}
\usepackage{algorithmic}
\usepackage{graphicx}
\usepackage{textcomp}
\usepackage{xcolor}
\def\BibTeX{{\rm B\kern-.05em{\sc i\kern-.025em b}\kern-.08em
    T\kern-.1667em\lower.7ex\hbox{E}\kern-.125emX}}
\begin{document}

\title{Analyze and Development System with Multiple Biometric Identification\\
}

\author{\IEEEauthorblockN{ Sher Dadakhanov}
\IEEEauthorblockA{\textit{Faculty of Computer Science} \\
\textit{Suleyman Demirel University}\\
Kazakhstan, Almaty region, city Kaskelen \\
181107027@stu.sdu.edu.kz\\
Kazakhstan, Almaty region, city Kaskelen, street Abylaikhan 1/1}}
\maketitle

\begin{abstract}
Cause of a rapid increase in technological development, increasing identity theft, consumer fraud, the threat to personal data is also increasing every day. Methods developed earlier to ensure personal
the information from the thefts was not effective and safe. Biometrics were introduced when it was needed
technology for more efficient security of personal information. Old-fashioned traditional approaches like Personal identification number( PIN), passwords, keys, login ID can be forgotten, stolen or lost. In biometric authentication system, user may not remember any passwords or carry any keys. As people they recognize each other by the physical appearance and behavioral characteristics that biometric systems use physical characteristics, such as fingerprints, facial recognition, voice recognition, in order to distinguish between the actual user and scammer. In order to increase safety in 2005, biometric identification methods were developed government and business sectors, but today it has reached almost all private sectors as Banking, Finance, home security and protection, healthcare, business security and security etc. Since
biometric samples and templates of a biometric system having one biometric character to detect and the user can be replaced and duplicated, the new idea of merging multiple biometric identification technologies has so-called multimodal biometric recognition systems have been introduced that use two or more biometric data characteristics of the individual that can be identified as a real user or not.
\end{abstract}

\begin{IEEEkeywords}
Fingerprint, face recognition, voice recognition, biometric identification, security, accuracy.
\end{IEEEkeywords}

\section{Introduction}
In today's, security is the main problem and everybody will be looking for reliable ways to ensure safety. Old ways of securing our homes and things used key and lock systems, which is quite inefficient keys can be duplicated. Later, the next level of protection was provided by modern technologies with PIN and
password. Modern systems are quite effective, but they can also be hacked and gain access. To achieve with high accuracy and reliability, a biometric identification technology has been introduced that uses physical or behavioral characteristics to verify the individual. The main reason for choosing biometric identification The technology above the traditional identification method is that traditional identification techniques are vulnerable
security, unreliable and uncertain. Most developed biometric systems use one biometric technology for biomass identification and verification. These unimodal systems are unable to meet some of the challenges such as enrollment large population, high performance etc. Later advanced biometric identification technology with better
Performance and a high degree of accuracy, called a multimodal biometric recognition system, are introduced. Multimodal systems are capable of using two or more physical or behavioral characteristics of a person write, verify and provide a decision. Multimodal biometric systems also combine two or more identification technologies, such as systems, fingerprint systems and face recognition, with more sensors and provides decisions by combining results from each subsystem. When using multimodal biometric systems, if one sensor or the system fails to carry out identification, other systems can be used to make decisions. Multimodal systems can be used in various fields, for example in homes for the biometric door locks to avoid theft, hospitals to record details of patient in software companies and educational institutions, to ensure participation in the banks for safe
and secure payment processing, biometric alarm systems in case of data security in different sectors. Multimodal systems are more convenient and overcome all limitations of unimodal systems with increased security, high accuracy and liveliness.

\section{General Statement of The Multimodal Authentication Problem}
The task of multimodal authentication consists of several steps:\cite{b1}\newline
1.	Get results from individual systems.\newline
2.	To normalize these results.\newline
3.	Perform the merger of normalized components.

\section{Related works}
In recent years, a lot of work has been done on different types of biometric identification techniques and their work, advantages and functioning of multimodal biometrics compared to unimodal and traditional approaches.\cite{b2} Biometrics refers to the automatic authentication of a person based on his physiological and behavioral property. The use of biometrics as a reliable method of verification occurs in almost all EU countries government and private sector. Some limitations of unimodal biometric systems may be reduced by the use of multimodal biometric systems that integrate information at different levels to improve performance.\cite{b3} In this the paper proposed a new multimodal biometric recognition system with a fusion of fingerprints and irises recognition to achieve higher accuracy. The main goal of this work is to show that the merger of unrelated, independent identification techniques achieve better accuracy than any unimodal biometric systems. Output scores from two different systems merge into a single score at the score level using three standardizations methods and four Fusion approaches. They also used standardization steps to transform unrelated, different scores to the common scale of the two techniques. In this article, they told about the types of mergers level in biometric systems. Sensor level, function level, score level are types of Fusion levels. First step on the score level is the normalization of the score, when the scores are transformed into a common format. Step two the Fusion level has two categories, such as classification and combination. Classification approach classifies decisions to a cheater or real. Combined approach combines score multiple decisions into one comparative score. The accuracy achieved by this work is more than 60 percentage.\cite{b4} By this article, I understand that a merger at the rank level can be a fairly new merger access. Fusion is carried out at the level of rank, when the output of each biometric system was a subset of possible matches sorted by declining self-esteem. The intention of the merger at rank level was to combine the output individual biometric subsystems to achieve greater accuracy. They used facial recognition, ear signature recognition and recognition systems and processed ones that use Eigen and Fisher face projection projection for registration and identification of the face, projection of own ears for registration and identification of ears, and Self-Signature projection for registration and signature identification. So use the assignment function based on in the ranks of the face, ear and signature there was a merger at the level of rank, and the final exit from them was obtained. From their by experiment, they came to the conclusion that the fishing face works more efficiently than the technique of its own face and of the three biometric face recognition systems, it worked or worked better than its own ear and your own signature.\cite{b5} In this article, they considered the extraction and evaluation of fingerprints, as well as the extraction of the veins of the fingers and conversion to the knowledge of a person. They also combined two biometric recognition systems using Enhanced partial discrete Fourier transformation (EP-DFT), since the DfT Matrix has its own characteristics, such as orthogonality. Like The DfT Matrix shortens the time for its multiplication by The Matrix, so that it becomes a partial discrete Fourier transformation compact and computationally efficient. They considered three different merger options where they have considered a set of fingerprint functions based on markants and a set of fingerprint functions based on an image using an element level Fusion Strategy. Their analysis concluded that EP-DFT significantly increases system security compared to EP-DFT original P-DFT. They also concluded that it is necessary to design a good, effective irreversible transformation functions for cancellable multi-biometric systems.

\subsection{EIGENFACES METHOD}
The basis of the eigenfaces method is the Principal
Component Analysis (PCA). Eigenfaces and PCA have been
used by Sirovich and Kirby to represent the face images
efficiently.\cite{b6} They have started with a group of original
face images, and calculated the best vector system for image
compression. Then Turk and Pentland applied the Eigenfaces
to face recognition problem. The Principal Component
Analysis is a method of projection to a subspace and is
broadly used in pattern recognition. An objective of PCA is
the replacement of correlated vectors of large dimensions with
the uncorrelated vectors of smaller dimensions. Another
objective is to calculate a basis for the data set. Main advantages of the PCA are its low sensitivity to noise, the reduction of the requirements of the memory and the capacity, and the increase in the efficiency due to the operation in a space of smaller dimensions.

\subsection{VIOLA-JONES}
This method is highly effective for search objects in images and video sequences in real time.\cite{b7} This detector has an extremely low probability of false detection of the face. The method works well and detects facial features even when observing an object at a slight angle, up to about 30°. Recognition accuracy using this method can reach values over 90 percentages, which is a very good result. When the angle of inclination is more than 30°, the probability of detecting a face drops sharply. The indicated feature of the method does not allow in a standard implementation to detect a person’s face, turned at an arbitrary angle, which greatly complicates or makes it impossible to use the algorithm in modern production systems, taking into account their growing needs.

\section{PROPOSED APPROACH}\label{SCM}
Individual or unimodal biometric systems are not enough for a person to effectively recognize a person, so I suggest multi biometric system for authentication to achieve better safety and reliability. I focus in particular on the use and degree of accuracy achieved by multibiometric recognition systems in domestic security and various
area. I will explore more about biometric recognition systems in different areas.

\subsection{METHODOLOGY}  
The following steps are procedures for identifying an individual.

\begin{figure}[!ht]
 
\centerline{\includegraphics[width=0.4\textwidth]{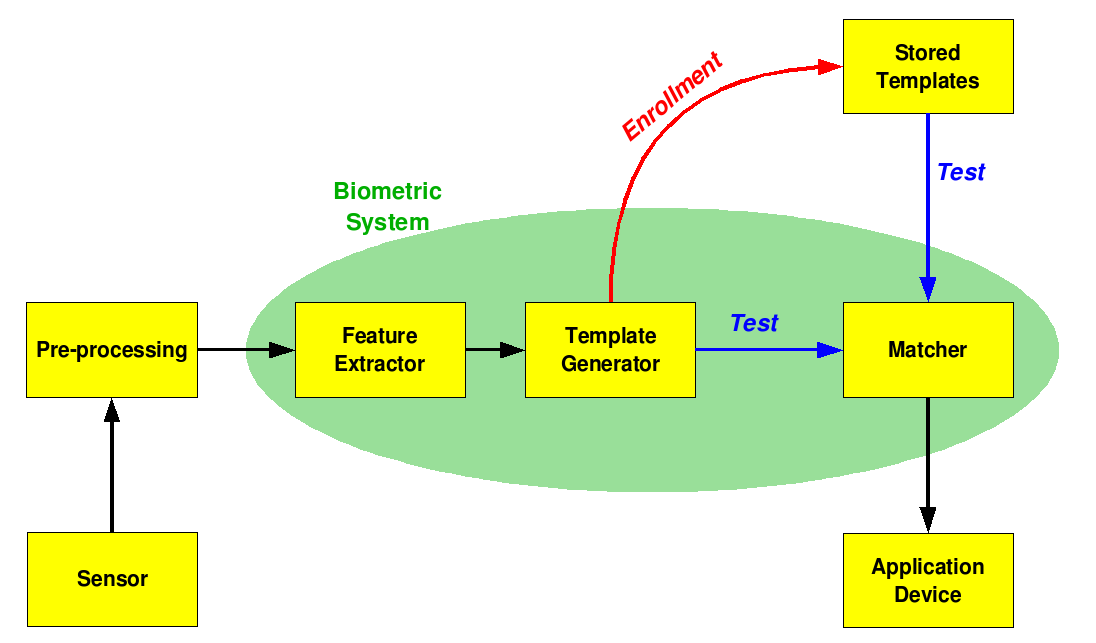}}
\caption{Simple diagram showing the main logical block of a biometric system}
\label{optaplanner}
\end{figure}
\begin{enumerate}
    \item \textbf{Enrollment of the biometric samples of a candidate}
    Enrollment is when a candidate first uses a biometric system, sample a biometric character as a fingerprint or face recognition is captured and stored in a database for later comparison.
    \item \textbf{Obtain/Acquire Live Samples from Candidate}
    First, get a biometric sample from the user using sensors.
    \item \textbf{Extract feature}
    Process and extract prominent features from the captured sample.
    \item \textbf{Comparison of samples}
    Process and extract prominent features from the captured sample....
    \item \textbf{Display decision}
    When the input samples coincide with the registered samples, it accepts the candidate or refuses.
\end{enumerate}

\section{FLOW DIAGRAM}
The below is the flow of working procedure of multi biometric recognition systems.
\begin{figure}[!ht]
\centerline{\includegraphics[width=0.3\textwidth]{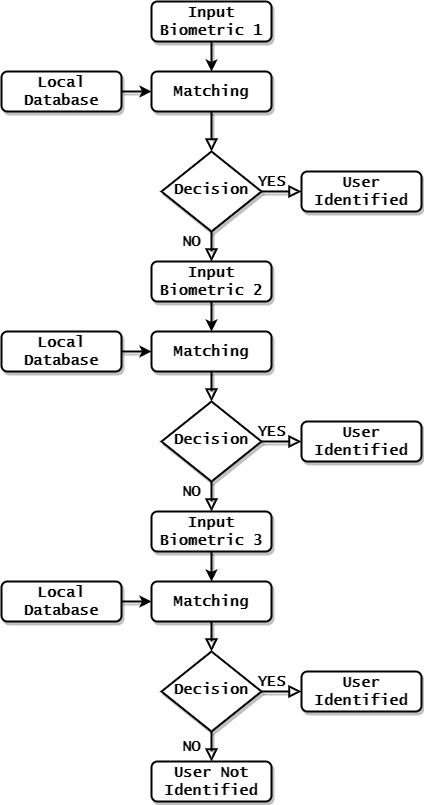}}
\caption{Flow Diagram of Multimodal Biometric Identification Systems}
\label{optaplanner}
\end{figure}

\begin{itemize}
    \item Initially, the sample biometric function is captured from the user when the user wants access. Biometric features can be fingerprints, voice recognition, face recognition.
    \item The captured biometric data is pre-processed, which includes the removal of unwanted data, noise and a prominent feature is obtained on which a person is to be recognized
    \item The extracted function is compared to registered samples that are stored in the database both samples have similarities.
    \item The logged sample and the captured sample are compared at different angles using some defined algorithms and corresponding accuracy are displayed.
    \item Based on the degree of accuracy, the decision is made as a real user or fraudster
    \item If one biometric system provides 60 percent accuracy, which is not sufficient for any decision other a system with a different biometric character is used in the same procedure.
    \item  The final results obtained from all subsystems are scaled and transformed into a normal format; and the decision appears on the screen.
\end{itemize}

Nowadays, biometrics have become a new trend in public and private workspace, where data or any other personal security is a major concern. Most nations move away from the traditional way to secure
houses of technology road. Instead of the lock and key system, more biometric door locks are used, which is a combined system of two or more biometric identification techniques for verification and verification. The biometric door will have a latch with fingerprint recognition or palm print recognition in order to identify the homeowner. And were introduced videos, which offer an easy way to find out who's at the door, without get close to the door. Bells with video can be connected to smartphones via Wi-Fi and the user can get alert when someone approaches the door. Video Bell records video when the bell is pressed and it offers two-way audio communication that allows you to communicate with the visitor from anywhere through your phone. Other features include facial recognition technology, which identifies visitors by name, motion sensing technology that knows the difference between humans, cars and animals.  Biometrics is also used in many organizations and educational institutions to record employees participation. The Multi biometric attendance system provides employees with flexibility in recording their presence; and working time at the right time interval. The main advantage of using multimodal biometric systems is fast identification and authentication. If one of the modalities fails to determine and submit a decision, the user can choose different modality or based on the convenience of the user, he can choose a modality to gain access. Biometrics are
it is also used in financial sectors such as banks, ATMs, insurance companies for Secured Transactions. Financial sectors are in terms of transactions very sensitive, because spoofing can lose a huge amount of money. To ensure secured transactions, financial sectors adopt biometric identification techniques identify account holders and also for KYC. Scanning the veins and scanning the iris are popular today where the user does it it does not require contact with the sensor and is more reliable and faster than other biometric technologies. Healthcare the Centers also use biometrics to identify the right patient, to provide a medical history and patient treatment plan.

\section{Conclusion}
In this digital era, most existing biometric systems use information from a single biological function verification and identification. The captured sample is due to the vulnerability of the biometric sensor to noisy or bad data biometric functions could become fake and it will be possible to reject the Registered User incorrectly
and falsely accept the fake user. In the case of a fingerprint recognition system, scanners may not recognize dirty or injured fingers also underdeveloped fingerprint combinations in the case of young children and faded fingerprint combinations in the case of the elderly. Where, as in the case of a facial recognition system, sensors may not be able to detect identical twins, whose appearance will be almost similar. Also, in the case of voice recognition systems, a person can imitate the voice of another person and gain access to their personal information. In all of the above cases due to an erroneous decision from the scanning devices gives unauthorized access and leads to fraudulent activity. In order to overcome these problems, a biometric system with a merger of two or more biological characteristics and sensors called multi-biometric systems are used. The main advantage of using multimodal biometric recognition there is an increased accuracy of verification. Thus, the use of multi-biometric recognition systems helps us in identifying and more precisely, in comparison with the use of a single biometric recognition system. I learned about different biometric recognition systems in order to use the necessary multimodal method biometrics for smart homes to have safe and reliable access to our homes. I can do this multimodal biometrics are more reliable and accurate compared to unimodal biometric recognition systems.

\vspace{12pt}


\begin{thebibliography}{00}
\bibitem{b1} A.RossandA.K.Jain. InformationFusioninBiometrics
//Proc. of AVBPA : conference. — June 2001. — P. 354—359.

\bibitem{b2}  A.H. Mir, S. Rubab, Z. A. Jhat, ―Biometrics Verification: A Literature Survey‖, International Journal of
Computing and ICT Research

\bibitem{b3} Kamer Vishi, Sule Yildirim Yayilgan, ―Multimodal Biometric Authentication using Fingerprint and Iris
Recognition in Identity Management‖, 2013 Ninth International Conference on Intelligent Information Hiding
and Multimedia Signal Processing, Faculty of Computer Science and Media Technology, Dept. of Information
Security Gjøvik University College

\bibitem{b4}   Md. Maruf Monwar, ―Multimodal Biometric System Using Rank-Level Fusion Approach‖, IEEE Transactions on Systems, Man, and Cybernetics—Part B: Cybernetics, Vol. 39, No. 4, August 2009, Student Member, IEEE, and Marina L. Gavrilova, Member, IEEE.

\bibitem{b5} Wencheng Yang, Song Wang, Jiankun Hu, Guanglou Zheng, Craig Valli, ―A Fingerprint and Finger-vein Based Cancelable Multi-Biometric System‖, Pattern Recognition Volume 78, June 2018, Pages 242-251, 2018.

\bibitem{b6} M.A. Turk and A.P. Pentland, “Face Recognizing Using Eigenfaces”, Proc. IEEE, 1991, 586-591.

\bibitem{b7} Viola, P. Robust realtime face detection / P. Viola
// International Journal of Computer Vision. – 2004.

\end{thebibliography}
\end{document}